\documentclass[conference]{IEEEtran}

\IEEEoverridecommandlockouts                              %

\usepackage{amsmath,amssymb,amsfonts}
\usepackage{subfig}
\usepackage{graphicx}
\usepackage[export]{adjustbox}
\usepackage{array}
\usepackage{booktabs}
\usepackage{color,colortbl}
\usepackage{tikz,pgf,pgfplots}
\usepackage{tikzscale}
\usepackage{bibentry}
\usepackage{blindtext} 
\usepackage{makecell}
\usepackage{comment}
\usepackage{url}
\usepackage{mathtools}

\usepackage{xcolor}

\definecolor{mycolor1}{rgb}{0.8,0,0}
\definecolor{mycolor2}{rgb}{0,.5,0}
\definecolor{mycolor3}{rgb}{0,0,0.8}
\definecolor{mycolor4}{rgb}{0,.7,.7}
\definecolor{mycolor5}{rgb}{.4,.4,.4}

\title{\LARGE \bf
Digital Twins Below the Surface: Enhancing \\ Underwater Teleoperation
}

\author{Favour O. Adetunji$^{1}$, Niamh Ellis$^{2}$, Maria Koskinopoulou$^{1}$, Ignacio Carlucho$^{1}$, Yvan R. Petillot$^{1}$ \\ %
$^{1}$School of Engineering and Physical Sciences, Heriot-Watt University, Edinburgh, UK. \\
$^{2}$School of Mathematics and Computer Sciences, Heriot-Watt University, Edinburgh, UK \\ 
\{foa2001, nmfe2000, m.koskinopoulou, ignacio.carlucho, y.r.petillot\}@hw.ac.uk \\
\thanks{*This work was supported by EPSRC grant code EP/X024806/1,  Fugro Ltd, and by the Robotics \& Autonomous Systems CDT.}%
}

\begin{document}

\maketitle

\begin{abstract}
Subsea exploration, inspection, and intervention operations heavily rely on remotely operated vehicles (ROVs). However, the inherent complexity of the underwater environment presents significant challenges to the operators of these vehicles. 
This paper delves into the challenges associated with navigation and maneuvering tasks in the teleoperation of ROVs, such as reduced situational awareness and heightened teleoperator workload.
To address these challenges, we introduce an underwater Digital Twin (DT) system designed to enhance underwater teleoperation, enable autonomous navigation, support system monitoring, and facilitate system testing
through simulation. 
Our approach involves a dynamic representation of the underwater robot and its environment using desktop virtual reality, as well as the integration of mapping, localization, path planning and simulation capabilities within the DT system. 
Our research demonstrates the system's adaptability, versatility and feasibility, highlighting significant challenges and, in turn, improving the teleoperators' situational awareness and reducing their workload.
\end{abstract}

\begin{IEEEkeywords}
Autonomous Underwater Vehicle, Digital-Twin, Simulator
\end{IEEEkeywords}

\section{INTRODUCTION}
Significant advancements in unmanned underwater vehicles (UUVs) have greatly expanded their applications and generated substantial research interest. In previous years, notable strides in autonomy, swarm behaviour, perception, sensing, and durability have been made. Despite these developments, the dynamic and complex nature of the marine environment still presents numerous challenges regarding the use and control of such vehicles.
One of these challenges involves the teleoperation of underwater remotely operated vehicles (ROVs), as factors such as poor visibility, reduced spatial awareness, increased teleoperator workload~\cite{geoffrey}, signal latency, and environmental disparities~\cite{tanwani} can occur during ROV teleoperation. 
These factors stem from various sources such as the nature of the underwater environment, the type of sensors used, communications channels, and the design of the teleoperation interface. 

Particularly, teleoperation interfaces can significantly affect the success of the underwater operation as they are the means by which the operator interacts with and controls the ROV.
Conventional underwater teleoperation interfaces typically rely on single or multiple video feedback displayed on screens~\cite{MONIRUZZAMAN}. However, this approach has inherent limitations, resulting in reduced spatial awareness for the teleoperator and increased workload due to the limited views of the robot's environment and state.  
In addition, with the trend of ground control of offshore assets~\cite{birk}, teleoperation interfaces are enhanced by transmitting live video streams and sensor data over long distances with low latency to cloud infrastructure or satellites while providing the operator with enough information and level of immersion.

\begin{figure}
    \centering
    \includegraphics[width=0.35\textwidth,valign=b]{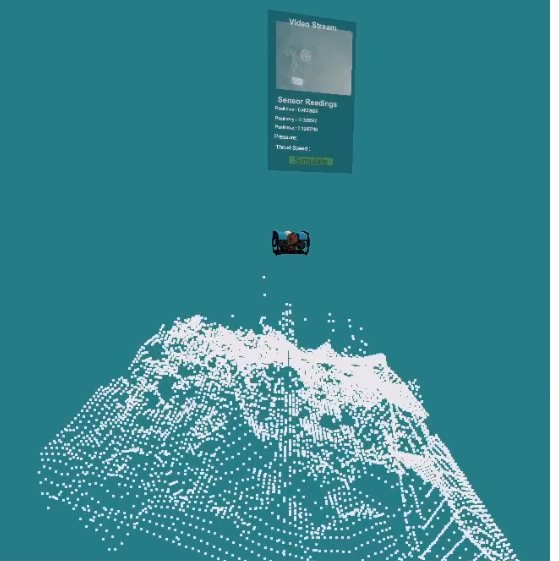}
    \caption{ DT interface featuring UUV visualization, point cloud seafloor recreation, and sensor data display.}
    \label{fig:DT_interface}
\end{figure}

Recent teleoperation interfaces employ virtual reality (VR)~\cite{elor}, to immerse the operator with the unique advantage of a three-dimensional environment that closely replicates the actual underwater scene, thereby helping to alleviate perceptual and workload issues~\cite{xia}. Extensions of VR for underwater teleoperation involve incorporating technologies such as mixed reality (MR), augmented reality (AR), and digital twin (DT) technology~\cite{DOMINGUES}~\cite{moreira} to achieve better teleoperator experience and UUV control, real-time monitoring, and training capabilities among others. %
In particular, the integration of DT technology with underwater teleoperation systems has succeeded in increasing the operator experience by providing additional context and information, real-time monitoring and analysis, simulation and training capabilities, as well as remote monitoring and control, ultimately improving the efficiency and effectiveness of underwater remote control systems. 
However, VR-based underwater teleoperation systems still face challenges, as issues such as the dynamic representation of the environment and the robot, the level of telepresence, and the realization of robust underwater VR systems have not yet been fully studied and resolved.

While there have been different definitions used for digital twins in the literature~\cite{mazumder_towards_2023}, in this work, the definition is based on~\cite{van_digital_2023} which defines the digital twin as initially being composed of three components: i) the 
digital component, ii) the physical component and, iii) the communication between the two.   
In this work, we present an underwater DT system that uses desktop virtual reality technology to dynamically represent the robot (physical component) and its environment, incorporating teleoperation, simulation and autonomous navigation functionalities. 
Our research presents a robust and practical DT system, marking a significant contribution to the under-explored realm of underwater digital twins.
The presented DT system can be integrated with ROS to control any type of underwater vehicle. 
The DT allows the operator to achieve an enhanced third-dimensional view of the environment with important sensor information, which facilitates teleoperation and monitoring. A view of the final interface of the digital component can be seen in Figure~\ref{fig:DT_interface}.
Furthermore, the DT system can be used for 
testing of system design and control algorithms through its simulation interface.  

The rest of the paper is organized as follows: first, related works are discussed in section \ref{RelatedWorks}. Subsequently, a brief background context on the physical component is presented in \ref{Background}, followed by a description of our system in section \ref{methodology}. Next, the results obtained with the designed DT system are discussed in section \ref{results}. A brief discussion of the future of the research area is provided in section \ref{Discussion}. Finally, the paper concludes with a conclusion and potential areas of future work (section \ref{conclusions}). 

\section{RELATED WORKS}\label{RelatedWorks}

Digital twins have been used in different industries such as aerospace, manufacturing and robotics~\cite{phanden_review_2021}. Furthermore, they have been used in different domains of robotics research, such as human-robot interaction and space robotics~\cite{mazumder_towards_2023}. Digital Twins have also been proposed for different uses in different areas of the marine industry such as in the ship-building industry~\cite{MarineIndustry}.  
The application of DT systems in underwater robotics can be challenging due to factors such as difficulty in high-fidelity modelling, communication breakdowns and interference issues arising from the inherently complex nature of subsea environments.
Recently, there has been a notable surge in research efforts concentrating on Digital Twin (DT) applications in the underwater domain, leading to diverse implementations of DT systems in this domain.
In~\cite{van_digital_2023} three different use cases of digital twins in various tasks in the domain are discussed. The first of these is in a launch and recovery scenario using a support vessel. The digital twin here is used to predict collisions between the UUV and the support vessel. The second use is in the control of a number of Autonomous Underwater vehicles (AUVs) working collaboratively. Here, the DT is used in predicting failures and an estimation of the remaining useful life (RUL) of the system. This then allows for tasks to be allocated in such a way as to allow the mission to be completed. The third case looks at the development of a trust-aware system. In this case, the trust-aware DT can be used to significantly decrease the time required for decision-making by taking the place of experts.

A DT system has been implemented for a sensor network in~\cite{barbie_developing_2022}. In this work, the designed system is tested with a marine deployment which allows the two systems (physical and digital) to be synchronised. 
DT systems have also been used in creating a simulation of the underwater environment for the development of a path planning algorithm for an Autonomous underwater glider (AUG)~\cite{yang_digital_2022}. Here the use of DTs aids in the environment being more representative of a true underwater environment compared to environments that are designed using mathematical methods. Unlike our work, this system exploits reinforcement learning and edge computing.   
It has also been suggested in~\cite{zuba_pairing_2022} that the use of DTs in the underwater domain does not require a communication system that is free of breakdowns and has a high data rate. And that the predictive ability of the DT system can be utilised to allow it to behave as though communication has not broken down. It is interesting to note that in~\cite{van_digital_2023} the issues of implementing wireless communication in the underwater domain are presented as a reason against the use of DTs in underwater robotics.  

In~\cite{Shanshan_2023}, researchers utilized DT technology for the development of a blended-wing-body underwater glider (BWBUG) vehicle. They adopted a methodology that involved the use of a high-fidelity simulation to validate the developmental and design stages of the BWBUG vehicle. The use of DT technology for underwater inspections is seen in~\cite{Kleiser_2020}. Here, the researchers presented a DT system designed for inspecting underwater structures like pipes. Their research methodology involved reconstructing the overall profile of an underwater inspected object in a virtual environment using elements like point clouds. The use of DT technology for underwater teleoperation can be seen in~\cite{xia}, here the authors developed a multi-sensory system with fluid flow simulation capabilities for assisting the teleoperator in navigating challenging underwater environments by providing insights into fluid dynamics.

In general, DT technology has significant potential to enhance UUVs operations. However, there is still a need for further research to integrate this technology into real-world underwater robotics applications.

\section{Background}\label{Background}

As previously stated, our definition of DT is based on~\cite{van_digital_2023}, which defines the digital twin as composed of the physical and digital components, as well as the communication between the two. In this section, we describe the basic elements constituting the physical component.  

\subsection{SLAM}

Simultaneous localisation and mapping (SLAM) must be performed. SLAM allows for both the robot's position and a map of the environment to be determined at the same time~\cite{SLAMoverview}. SLAM can be expressed mathematically as follows:
\begin{equation}
    \begin{dcases}
     x_{k}=f(x_{k},u_{k},w_{k})\\
      z_{k,j}=h(y_{i},x_{k},v_{k,j})
    \end{dcases}
\end{equation}
With $k$ used to denote a point in time, $x$ the robot's position, $u$ the robot's input and $w$ the noise. While $z_{k,j}$ is the observation data at the time instance $k$ for waypoint $j$. Additionally $y_{j}$ denotes the waypoint $j$ and $v_{k,j}$ is the noise at this waypoint at time $k$. Visual SLAM is often used in underwater robotics. This is because GPS cannot be used in the underwater environment ~\cite{SLAMoverview} and other forms of sensors can be more expensive and suffer adverse effects underwater ~\cite{ZHANG_Visual_2022}. Visual SLAM has multiple components such as front-end, back-end, loop-closure detection, initialization and mapping ~\cite{SLAMoverview}. An important part of this is the loop closure detection which is responsible for detecting when the ROV has reached a location that it has previously visited. This decreases the accumulated errors and improves the estimation of poses. 

\subsection{Path Planning}

As mentioned previously part of the designed system involves a path-planning algorithm. Path planning is an important part of the AUV's navigation process and involves designing a path from a starting position to a final goal position~\cite{chen_research_2021}. In the underwater environment path planning is quite challenging due to the variability of the environment and limited sensor information. Path planning methods can be classified as global or local ~\cite{Yao_Review_2019}. In global path planning the path is planned based on previously obtained information while in local path planning the path is planned based on the current sensor readings. Some examples of global path planning methods include the Rapidly Exploring Random Tree (RRT)  algorithm and the genetic algorithm. Examples of local path planning methods are the artificial potential field method and fuzzy logic algorithms.  Path planning has three main steps, environment modelling, positioning and path planning ~\cite{chen_research_2021}. The environment modelling stage is when the environment around the AUV is mapped into a model that can be used by a computer. In the positioning stage, the AUV's position is obtained with the path finally being planned in the path planning stage. In the environment modelling stage, the environment is represented via some method such as the grid method or the Voronoi diagram method. There are different ways of path planning such as artificial potential fields or random sampling methods. The second of these performs well in underwater environments while the first does not. Random sampling methods work by using random sampling to search, the RRT algorithm is one such method. It takes an initial point for a root and extends to create a random tree by randomly sampling nodes. Once the target point is in the tree a path between the initial and final points can be found. While capable of solving path planning problems in high-dimensional space and under complex constraints the path found may not be optimal. RRT*, an extension of RRT, on the other hand, finds the optimal path. The reason for this is that RRT searches for the nearest neighbour and RRT* checks multiple neighbouring nodes within a set radius. 

\section{Digital Twin}\label{methodology}
\begin{figure*}[h]
    \centering
    \includegraphics[trim={5cm 1.5cm 5cm 3cm},clip, width=0.7\textwidth]{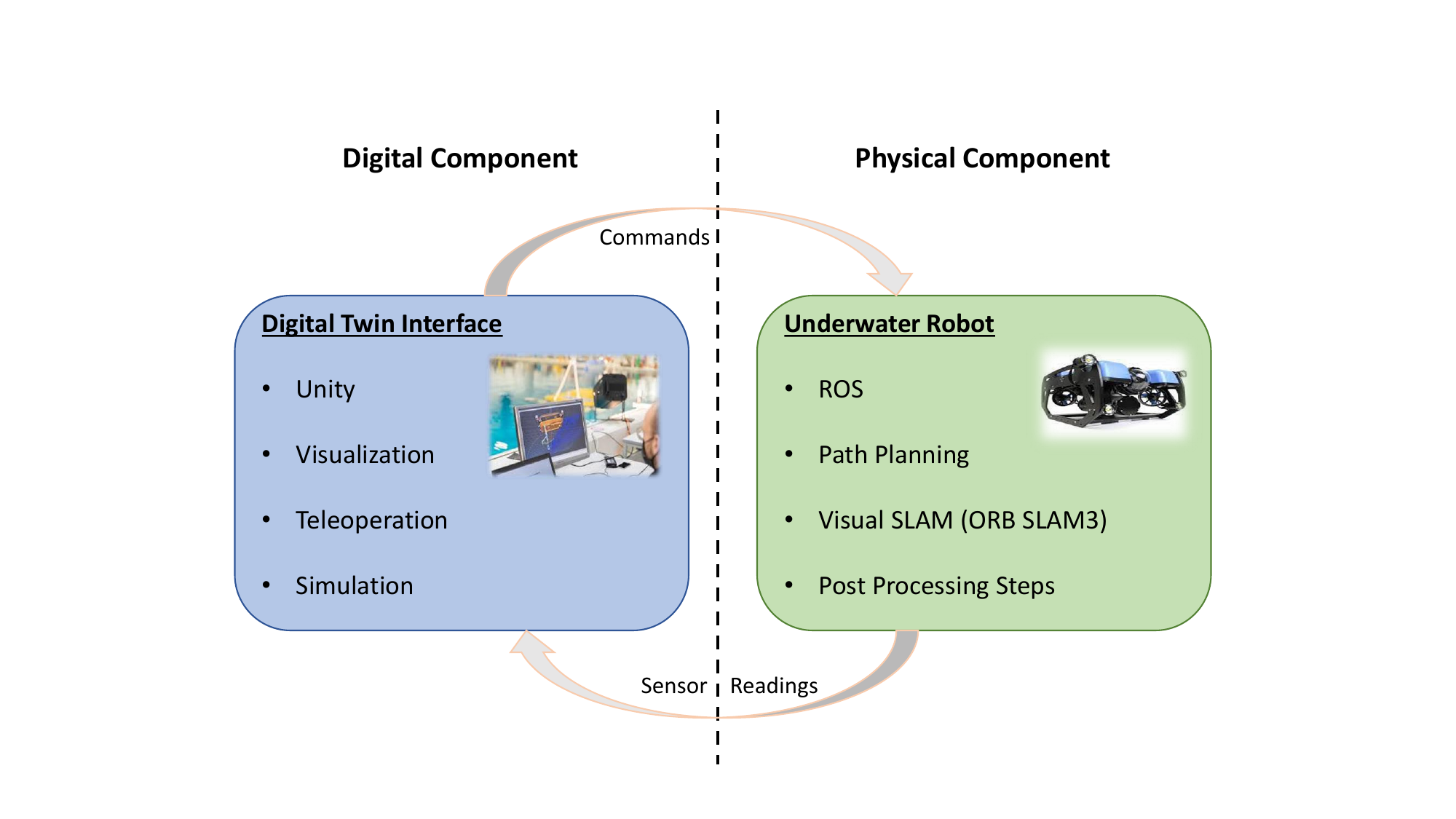}
    \caption{Digital Twin System Overview.}
    \label{fig:System_overview}
\end{figure*}

The architecture of the presented system consists of three fundamental components, illustrated in Figure~\ref{fig:System_overview}, aligning with the conventional Digital Twin model: the physical component, the digital component, and the bi-directional communication channel. These components collaborate synergistically, contributing to the overall functionality of the DT system.  In our DT system, our primary emphasis is on the BlueROV2 robot, which functions as the physical entity. The rest of this section details our system's methodology, elucidating the development process and intricacies of each component.

\subsection{Physical Component}
The physical component within the DT system represents the tangible and real-world entity that interacts with the digital component. The physical component in our system is the BlueROV2 robot~\cite{ScharffMarketReady}. The BlueROV2 utilizes the Robot Operating System (ROS), which enabled us to integrate the visual SLAM system, implement a  path planning algorithm, and execute post-processing steps (point cloud densification and map conversion).

We integrate the visual SLAM (ORB-SLAM3 \cite{campos}) system into the DT using a ROS-specific implementation. Within the physical component of the system, the visual SLAM process entails the meticulous reconstruction of the underwater environment, drawing information from each frame of the camera feed. As the algorithm operates, it transforms the camera input into point clouds, representing critical features observed by the camera within the underwater environment. Concurrently, the system engages in real-time monitoring of the robot's pose, a dynamic aspect that is continuously evolving as the robot moves through the environment. This monitoring is facilitated through the analysis of camera movements, providing insights into the spatial orientation and location of the robot within the environment. Considering ORB-SLAM3 is a sparse mapping algorithm, we extend our technique by employing a post-processing step for densifying the point cloud. We introduce an interpolation technique to transform these sparse point clouds into an intuitive surface representation of the environment within the physical component by taking the point clouds generated, interpolating between the coordinates points over a grid and subsequently, transmitting this data to the digital component of our system. 

The integration of the path planning features into the physical component of the system is achieved by utilizing an RRT* (Rapidly Exploring Random Tree Star) planner and an Octomap \cite{octomap} representation of the mapped environment. The RRT* algorithm, an extension of the widely used RRT algorithm, can efficiently explore and plan paths within high-dimensional spaces, which is well-suited for our particular use case. Incorporating the RRT* algorithm into our system involved utilizing the OMPL (Open Motion Planning Library) \cite{ompl} library through ROS. For compatibility reasons with the OMPL library, we implement a conversion process within our process workflow that transforms the ORB-SLAM3 map into an OctoMap representation. Additionally, we enhance collision-checking capabilities by integrating the flexible collision library (FCL) \cite{fcl}. This enables the efficient detection of potential collisions between the robot and various obstacles in the environment. The result of the path planner system is a set of waypoints. These waypoints represent the specific points in space that are necessary for the robot to follow so that it achieves the planned trajectory. 

Structurally, we execute the SLAM  and path planning computational tasks at the physical component of the system to minimize the complexity of the system as well as data sent to and from the digital and physical components.

\subsection{Digital Component}
The digital component of the DT system simply refers to the virtual representation or digital counterpart of the physical entity. The digital component in our system is responsible for the simulation, visualisation and teleoperation
of the DT system.

An integral element within our system's digital component is the DT interface. This interface plays a pivotal role, serving as the gateway to access the visualization, simulation, and control functionalities embedded within the DT system. It acts as a central hub through which the end user can interact with and manipulate the virtual representation of the physical entity.
 The DT interface is built upon the UWRobotics simulator~\cite{chaudhary}, a Unity-based robotics simulation platform. Within the interface, we create a three-dimensional interactive computer-generated environment (desktop virtual reality) that enables the manipulation and observation of the underwater robotic vehicle and its surroundings by utilizing the functionalities and capabilities of the Unity platform. The design of key elements in the DT interface incorporates concepts presented in \cite{BONINFONT}, including features such as multi-viewpoints or perspectives through simulated cameras, multi-sensory monitoring and point cloud generation for surface recreation. The interface is designed to achieve mirrored activities of the real robot through scripts in Unity (C\# scripts) tailored to process messages from the physical component of the system into visually understandable actions within the interface. Additionally, we design a keyboard-based input system and integrate it with the interface to facilitate manipulation of the virtual environment, such as viewpoint manipulation and to relay user actions to the physical component of the system such as control of the UUV (teleoperation).
The fundamental elements comprising the DT interface consist of game objects, including the robot model, the ocean environment, point cloud objects, simulated cameras, and text displays. These game objects encompass both 3D models of various entities and built-in elements within Unity. 

\begin{figure*}[t]
       \centering
          \subfloat[]{%
             \includegraphics[height=0.29\textwidth,valign=b]{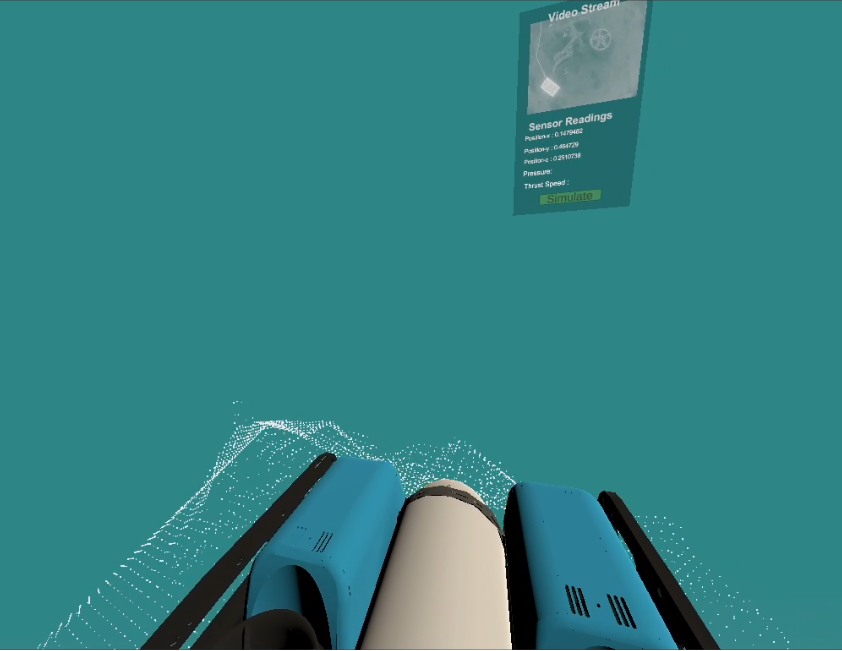}%
             \label{fig:DT_interface_3}%
          }\hfil
         \subfloat[]{%
             \includegraphics[height=0.29\textwidth,valign=b]{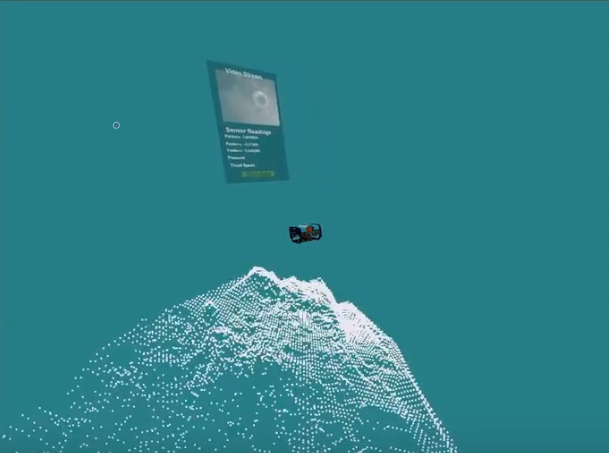}%
             \label{fig:DT_interface_4}%
          }
          \caption{ Digital Twin interface visualization: a) first person viewpoint; b) third person viewpoint.}
          \label{fig: Multiple_DT_interface}
\end{figure*}

Within the DT interface, we explore methods for dynamically representing entities, moving away from static representations in DT systems like pre-existing 3D models of environments. To do this, we dynamically represent the environment and the robot's pose within our system, through a visual SLAM implementation - ORB-SLAM3~\cite{campos}. 
In our system, this technique combines point clouds representing key environmental features with robot pose tracking. In this context, we engineer scripts (C\# scripts) that transform point cloud and robot pose messages originating from the physical component of the DT system into coherent tasks within the 3D environment embedded in the interface.
These scripts are responsible for orchestrating the generation and placement of point clouds within the environment, as well as controlling the movement of the robot's 3D model within this simulated space. The dynamic nature of the system within the DT interface is indicative of how the robot's movements within its environment are continuously replicated and updated within the interface.

Building upon the dynamic environment representation, we develop mechanisms for planning paths to designated goal points within our DT system. Our primary objective in this endeavour is to alleviate the teleoperator's workload through autonomous navigation capabilities within the DT by equipping the teleoperator with mechanisms to send path plan requests to the physical component. Thus, by enabling the planning and execution of simple paths within the previously traversed environment through the DT interface, we aim to eliminate repetitive actions, thereby alleviating potential increases in workload. This is accomplished within the DT interface by creating custom game objects.

For the integration of simulation capabilities into the digital component of the system, we depend on the UWRobotics simulator, which offers features like modelling hydrodynamics, vehicle kinematics, and dynamic properties in underwater environments. We integrate the simulator into the system to primarily facilitate teleoperation simulation in our system. However, the simulation interface also serves as a versatile platform for testing control algorithms and simulating various system designs. In its role as a teleoperation simulator, the platform empowers a user to operate an underwater robot within a simulated environment that closely mirrors real-life operations. Here, we incorporate the BlueROV2 robot and its properties into the simulation system and establish an interface that facilitates seamless transitions to the UWRobotics simulation environment.

\subsection{Bi-directional Communication Channel}
The bidirectional communication channel denotes the connection that facilitates the exchange of data between the physical and digital components. For this, we utilize the ROS bridge suite and ROSbridgeLib software libraries. These packages facilitate a web socket connection over TCP/IP protocols for communication which is viable as it is capable of being used over a protected ethernet cable in an underwater environment. The communication network architecture between the physical component and the digital component mirrors the ROS publish-subscribe architecture. In this setup, a node can publish over a specific topic, and another node can subscribe to the published data over the same topic. This results in messages being published from the DT's digital component and subscribed to by the physical component, and vice versa.

In practice, a continuous stream of data is transmitted between both components as the robot navigates through the underwater environment. This ongoing transmission plays a crucial role in updating and synchronizing the digital counterpart, ensuring that it accurately reflects the real-time status and dynamics of the physical robot within the digital environment. Simultaneously, it guarantees precise communication of teleoperator commands to the robot. The facilitation of both these transmission processes is carried out through the communication channel.

\begin{figure*}[t]
       \centering
          \subfloat[]{%
             \includegraphics[height=0.37\textwidth,valign=b]{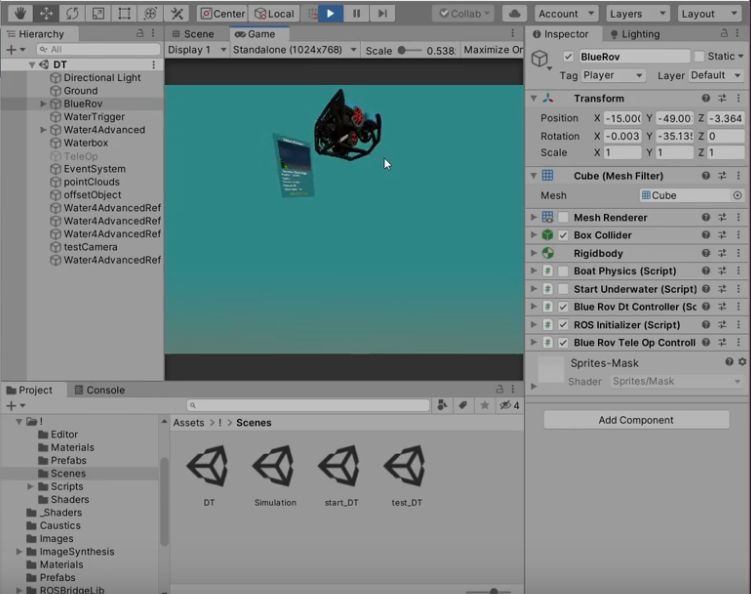}%
             \label{fig:teleop_left}%
          }\hfil
         \subfloat[]{%
             \includegraphics[height=0.37\textwidth,valign=b]{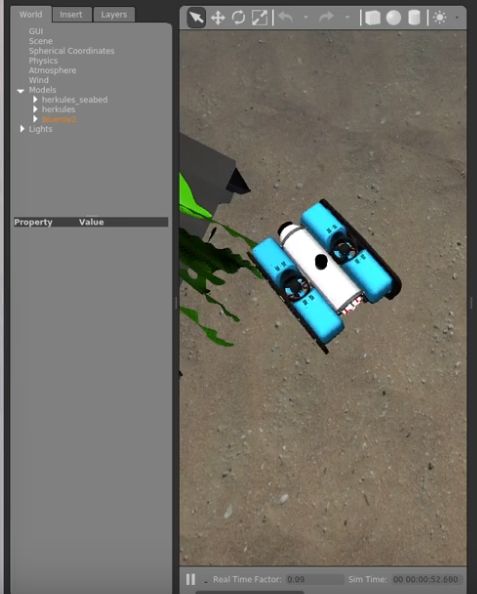}%
             \label{fig:teleop_right}%
          }
          \caption{Digital Twin teleoperation snapshots: a) Digital Twin interface; b) simulation environment.}
          \label{fig:teleop}
\end{figure*}

\section{SYSTEM INTEGRATION RESULTS}\label{results}

The DT system we have developed was subjected to testing using real-world data and in a simulation environment, with a primary emphasis on seafloor mapping. The results were obtained on a laptop with 16 gigabytes of RAM, an Intel Core i7 processor, and an Nvidia 3600 RTX graphics card. This system operates seamlessly on the Windows 11 operating system, supporting the Unity environment, and integrates a Windows Subsystem for Linux 2 (WSL2) running Ubuntu 18.04 and ROS Melodic.

Testing was performed on different elements of the DT system. The first set of tests was on the interface. This is done using real data (Rosbag datasets) from an ROV's teleoperation-led exploration of a harbour using a downward-facing monocular camera. The idea is that the DT system would re-create the environment and motion of the ROV during its exploration.
Figures \ref{fig:DT_interface_3} and  \ref{fig:DT_interface_4} illustrate the outcomes of the DT interface, when ROSbag datasets were executed to allow for replication of the operation of the real ROV during the harbour traversal. 
In all instances, the system displays an interpolated point cloud representation of the sea floor, along with the robot's movements and other sensory data being transmitted from the physical component of the system. The real robot's orientations are accurately recreated and by employing multiple simulated viewpoints the DT interface gives the operator an enhanced spatial awareness.

The teleoperation ability of the system was also tested by using an underwater simulator, the UUV simulator and a BlueROV2 ROS package for emulating a BlueROV2 robot in an underwater environment.  The testing of the developed controller was performed in a shipwreck environment contained in the UUV simulator, the ROV was controlled via the DT digital interface with the movements being accurately mirrored in the digital component of the DT system. Figure \ref{fig:teleop} illustrates the teleoperation of the robot through the DT interface and showcases the corresponding movement within the simulator. The system's success in transmitting the required controls was found to be at 100\% this indicated that the commands from the digital component were consistently received.

\begin{figure}
    \centering
    \includegraphics[height=4.8cm, valign=b]{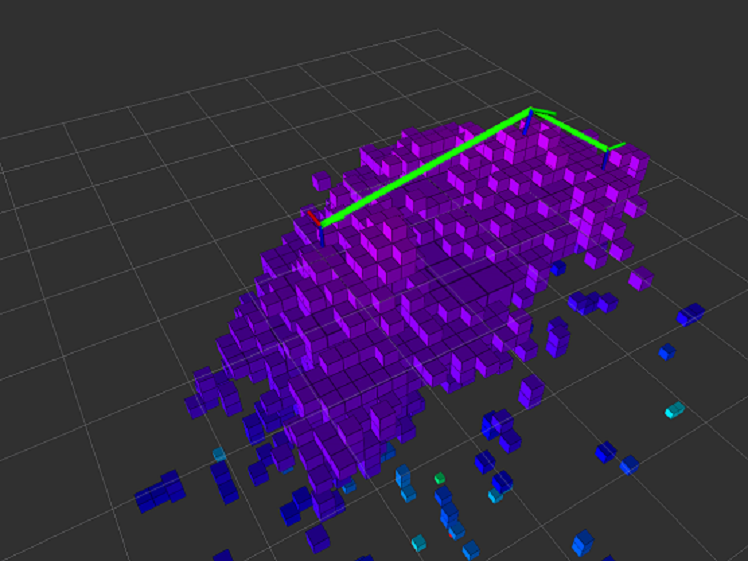}
    \caption{Octomap structure of mapped sea floor environment with the robot's planned path and waypoints (marked in green)}
    \label{fig:PP}
\end{figure}

Testing was also performed on the autonomous navigation system. Here, Rosbag datasets were used to replay the teleoperation process. As this was performed visual SLAM was used to localise the robot and map the environment based on the tracking of visual features. Following the completion of the route, the DT system utilises a function to convert the map and plan a path.  
In Figure \ref{fig:PP}, we provide visualization results of a planned path using the DT system, where the sea floor is depicted by using an Octomap structure and the waypoints of the planned path for the robot's movements are delineated by the green line. The path planning algorithm successfully created a path between start and end points for various scenarios. These included varying the end position of the path, paths that were prone to collisions based on the start and end points and also those that required navigation close to the sea floor. The RRT* algorithm performance was affected as the tests were performed under time conditions. The system had trouble generating paths when the path was in the last two states. However, the planned trajectories of simple paths i.e. those that were not near the sea floor or collision-prone were 100\% successful. The more complex paths involving start or end positions that were close to the sea floor were less successful in terms of planned trajectory however had a lower success rate. This success rate was nevertheless above 50\% under all time constraints.  

The issue of delay was noted during the design and testing of the system.
Particularly, this delay was observed during information transfer from the physical component to the digital component, as a larger amount of data was being transmitted compared to the reverse direction.  Testing was conducted to precisely observe the delay in the system by examining the expanded time during data transfer. The various ROS message types observed were those primarily used in teleoperation and dynamic environment representation which are the point cloud 2 message, sensor image message (camera raw image), pose stamped message (robot pose) and wrench message type (teleop command).

\begin{figure}
 \centering
    \includegraphics[height=6.2cm, valign=b]{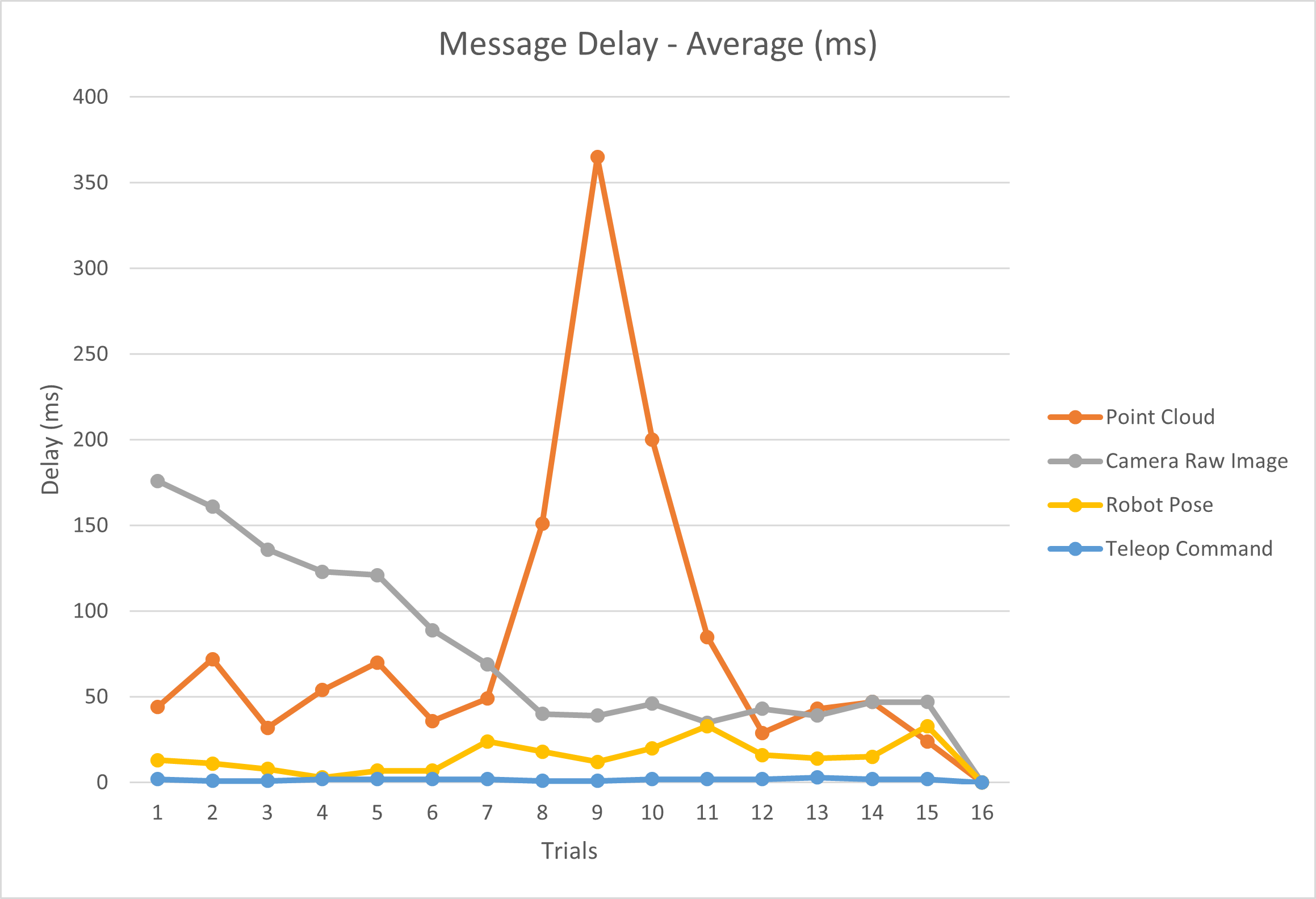}
    \caption{DT system analysis: Plot of different message types against delay (ms).}
    \label{fig:delay}
 \end{figure}

Figure \ref{fig:delay} %
illustrates the results of the methodical examination of the interactions among these message types employed in the DT system, shedding light on how they impact the system's performance. Here, we document and illustrate the expanded time between the transfer of these message types across 15 continuous timestamps in our system. Our analysis reveals a clear connection between bandwidth and latency within our system. Specifically, point cloud message types with the highest bandwidths also encountered the most significant delays while the reverse was the case for wrench message types with the lowest bandwidth.

\section{DISCUSSION}\label{Discussion}
The DT system developed here was designed to work with an underwater robot. Given the relatively new expansion of DT systems into the field a brief discussion of potential applications and challenges is now provided. 

As previously mentioned there are differing opinions about the potential difficulties communication may have over the system. It was suggested in~\cite{zuba_pairing_2022} that the predictive ability of the DT system could be employed to overcome this. Alternatively, it has been suggested that the challenges in implementing wireless communications in the underwater domain will be a large hurdle to implementing DTs in these types of systems. At present it is difficult to predict whether this will be an issue but the implementation of an underwater sensor network with a corresponding DT network ~\cite{barbie_developing_2022} is promising. 

Our system currently is designed to handle a single path planning situation but can be expanded for use in teleoperation systems. Given the increased points of view compared to the traditional camera view used by teleoperators the system is likely to give a huge benefit to the operators. In recent work on the teleoperation of ROVs VR has featured~\cite{elor} as have other methods of enhancement such as haptic feedback~\cite{xia_visual-haptic_2023} and gesture control~\cite{garcia_natural_2017}. While these methods likely could be beneficial there are some issues that our system has the potential to overcome. Such as motion sickness and discomfort wearing the device for the operators. Further involving ROV operators in the design of the system is likely to be beneficial to ensure that the system actually aids operators. It will also likely go a long way in their acceptance of the solution, which will be essential in getting any solution adopted. 
Operators are willing to lend their experience to research whether in the testing of a new system~\cite{elor} or discussing their role   ~\cite{louca_elicitation_2023, teigland_operator_2020}. In ~\cite{louca_elicitation_2023} teleoperators from various domains including underwater are surveyed to determine their feelings about their systems and under what conditions autonomy would be desired. When discussing the requirements for an operator to gain confidence in the system it is suggested that ROV operators must both be able to understand how the vehicle operates, its limitations and what steps they can take if things go wrong. This shows that from an operator's perspective, they would like to be involved with the design of the system as it improves their confidence in the system.

\section{CONCLUSION}\label{conclusions}
A DT system for the teleoperation of a ROV has been designed in this work. The designed system addresses two flaws in the teleoperation of an ROV; the spatial awareness and workload of an operator. Alongside the recreated dynamic environment created as a part of the DT system, autonomous navigation capabilities are also included. Most importantly the work here includes the creation of a dynamic representation of the environment and re-engineering of robot poses using ORB-SLAM3. Of equal importance, both a teleoperation input control system and a path-planning algorithm are included in the system. The currently designed system validates the use of DT systems with teleoperated underwater robots but can also be used as a foundation for further work.    

Possible avenues of future work are now discussed. As a first step the designed system needs to be implemented on an actual robotic system instead of in simulation this will allow for more testing. We believe that doing so will further validate our designed system. This will also open up avenues for more development such as the development of a motion planning system which was not possible with the current set-up of the system. 

Secondly, improving the point cloud densification process in the system can be beneficial, given that the current linear interpolation method falls short of accurately representing the environment. The inherent methodology of interpolating between coordinate point clouds may compromise the fidelity of the representation. To address this issue, a potential solution involves integrating a more advanced technique, such as a dense SLAM method or utilizing superior interpolation techniques. This enhancement aims to achieve a more accurate depiction of the environment in the generated point cloud.

\nocite{*}
\bibliographystyle{IEEEtran}
\bibliography{main}

\begin{thebibliography}{10}
\providecommand{\url}[1]{#1}
\csname url@samestyle\endcsname
\providecommand{\newblock}{\relax}
\providecommand{\bibinfo}[2]{#2}
\providecommand{\BIBentrySTDinterwordspacing}{\spaceskip=0pt\relax}
\providecommand{\BIBentryALTinterwordstretchfactor}{4}
\providecommand{\BIBentryALTinterwordspacing}{\spaceskip=\fontdimen2\font plus
\BIBentryALTinterwordstretchfactor\fontdimen3\font minus \fontdimen4\font\relax}
\providecommand{\BIBforeignlanguage}[2]{{%
\expandafter\ifx\csname l@#1\endcsname\relax
\typeout{** WARNING: IEEEtran.bst: No hyphenation pattern has been}%
\typeout{** loaded for the language `#1'. Using the pattern for}%
\typeout{** the default language instead.}%
\else
\language=\csname l@#1\endcsname
\fi
#2}}
\providecommand{\BIBdecl}{\relax}
\BIBdecl

\bibitem{geoffrey}
G.~Ho, N.~Pavlovic, R.~Arrabito, and R.~Abdalla, ``Human factors issues when operating unmanned underwater vehicles,'' \emph{Proceedings of the Human Factors and Ergonomics Society Annual Meeting}, vol.~55, 09 2011.

\bibitem{tanwani}
A.~K. Tanwani and S.~Calinon, ``A generative model for intention recognition and manipulation assistance in teleoperation,'' in \emph{2017 IEEE/RSJ International Conference on Intelligent Robots and Systems (IROS)}, 2017, pp. 43--50.

\bibitem{MONIRUZZAMAN}
M.~Moniruzzaman, A.~Rassau, D.~Chai, and S.~M.~S. Islam, ``Teleoperation methods and enhancement techniques for mobile robots: A comprehensive survey,'' \emph{Robotics and Autonomous Systems}, vol. 150, p. 103973, 2022.

\bibitem{birk}
A.~Birk \emph{et~al.}, ``Dexterous underwater manipulation from onshore locations: Streamlining efficiencies for remotely operated underwater vehicles,'' \emph{IEEE Robotics \& Automation Magazine}, vol.~25, no.~4, pp. 24--33, 2018.

\bibitem{elor}
A.~Elor \emph{et~al.}, ``Catching jellies in immersive virtual reality: A comparative teleoperation study of rovs in underwater capture tasks,'' in \emph{in proc. of the 27th ACM Symposium on VR Software and Technology}, ser. VRST '21.\hskip 1em plus 0.5em minus 0.4em\relax Association for Computing Machinery, 2021.

\bibitem{xia}
P.~Xia, F.~Xu, Z.~Song, S.~Li, and J.~Du, ``Sensory augmentation for subsea robot teleoperation,'' \emph{Computers in Industry}, vol. 145, p. 103836, 2023.

\bibitem{DOMINGUES}
C.~Domingues, M.~Essabbah, N.~Cheaib, S.~Otmane, and A.~Dinis, ``Human-robot-interfaces based on mixed reality for underwater robot teleoperation,'' \emph{IFAC Proceedings Volumes}, vol.~45, no.~27, pp. 212--215, 2012.

\bibitem{moreira}
M.~Laranjeira~Moreira, A.~Arnaubec, L.~Brignone, C.~Dune, and J.~Opderbecke, ``3d perception and augmented reality developments in underwater robotics for ocean sciences,'' \emph{Current Robotics Reports}, vol.~1, 09 2020.

\bibitem{mazumder_towards_2023}
\BIBentryALTinterwordspacing
A.~Mazumder, M.~F. Sahed, Z.~Tasneem, P.~Das, F.~R. Badal, M.~F. Ali, M.~H. Ahamed, S.~H. Abhi, S.~K. Sarker, S.~K. Das, M.~M. Hasan, M.~M. Islam, and M.~R. Islam, ``Towards next generation digital twin in robotics: {Trends}, scopes, challenges, and future,'' \emph{Heliyon}, vol.~9, no.~2, p. e13359, Feb. 2023. [Online]. Available: \url{https://www.sciencedirect.com/science/article/pii/S2405844023005662}
\BIBentrySTDinterwordspacing

\bibitem{van_digital_2023}
\BIBentryALTinterwordspacing
M.~Van, C.~Edwards, L.~Tran-Thanh, and M.~Bonney, ``\BIBforeignlanguage{en}{Digital {Twins} for {Marine} {Operations}: {From} {Surface} to {Deep} {Water}},'' EPSRC UK-RAS Network, {UKRAS} {White} {Papers}, Oct. 2023, edition: 1. [Online]. Available: \url{https://www.ukras.org.uk/publications/white-papers/digital-twins-for-marine-operations/}
\BIBentrySTDinterwordspacing

\bibitem{phanden_review_2021}
\BIBentryALTinterwordspacing
R.~K. Phanden, P.~Sharma, and A.~Dubey, ``A review on simulation in digital twin for aerospace, manufacturing and robotics,'' \emph{Materials Today: Proceedings}, vol.~38, pp. 174--178, Jan. 2021. [Online]. Available: \url{https://www.sciencedirect.com/science/article/pii/S2214785320349245}
\BIBentrySTDinterwordspacing

\bibitem{MarineIndustry}
\BIBentryALTinterwordspacing
Z.~Lv, H.~Lv, and M.~Fridenfalk, ``Digital twins in the marine industry,'' \emph{Electronics}, vol.~12, no.~9, 2023. [Online]. Available: \url{https://www.mdpi.com/2079-9292/12/9/2025}
\BIBentrySTDinterwordspacing

\bibitem{barbie_developing_2022}
A.~Barbie, N.~Pech, W.~Hasselbring, S.~Flögel, F.~Wenzhöfer, M.~Walter, E.~Shchekinova, M.~Busse, M.~Türk, M.~Hofbauer, and S.~Sommer, ``Developing an {Underwater} {Network} of {Ocean} {Observation} {Systems} {With} {Digital} {Twin} {Prototypes}—{A} {Field} {Report} {From} the {Baltic} {Sea},'' \emph{IEEE Internet Computing}, vol.~26, no.~3, pp. 33--42, May 2022.

\bibitem{yang_digital_2022}
\BIBentryALTinterwordspacing
J.~Yang, M.~Xi, J.~Wen, Y.~Li, and H.~H. Song, ``A digital twins enabled underwater intelligent internet vehicle path planning system via reinforcement learning and edge computing,'' \emph{Digital Communications and Networks}, May 2022. [Online]. Available: \url{https://www.sciencedirect.com/science/article/pii/S2352864822000967}
\BIBentrySTDinterwordspacing

\bibitem{zuba_pairing_2022}
M.~Zuba, ``Pairing {Digital} {Twins} with {Maritime} {Autonomy},'' in \emph{{OCEANS} 2022, {Hampton} {Roads}}, Oct. 2022, pp. 1--6, iSSN: 0197-7385.

\bibitem{Shanshan_2023}
\BIBentryALTinterwordspacing
S.~Hu, Q.~Liang, H.~Huang, and C.~Yang, ``Construction of a digital twin system for the blended-wing-body underwater glider,'' \emph{Ocean Engineering}, vol. 270, p. 113610, 2023. [Online]. Available: \url{https://www.sciencedirect.com/science/article/pii/S0029801822028931}
\BIBentrySTDinterwordspacing

\bibitem{Kleiser_2020}
D.~Kleiser and P.~Woock, ``Towards automated structural health monitoring for offshore wind piles,'' in \emph{Global Oceans 2020: Singapore – U.S. Gulf Coast}, 2020, pp. 1--5.

\bibitem{SLAMoverview}
\BIBentryALTinterwordspacing
X.~Wang, X.~Fan, P.~Shi, J.~Ni, and Z.~Zhou, ``An overview of key slam technologies for underwater scenes,'' \emph{Remote Sensing}, vol.~15, no.~10, 2023. [Online]. Available: \url{https://www.mdpi.com/2072-4292/15/10/2496}
\BIBentrySTDinterwordspacing

\bibitem{ZHANG_Visual_2022}
\BIBentryALTinterwordspacing
S.~Zhang, S.~Zhao, D.~An, J.~Liu, H.~Wang, Y.~Feng, D.~Li, and R.~Zhao, ``Visual slam for underwater vehicles: A survey,'' \emph{Computer Science Review}, vol.~46, p. 100510, 2022. [Online]. Available: \url{https://www.sciencedirect.com/science/article/pii/S1574013722000442}
\BIBentrySTDinterwordspacing

\bibitem{chen_research_2021}
\BIBentryALTinterwordspacing
Y.~Guo, H.~Liu, X.~Fan, and W.~Lyu, ``Research {Progress} of {Path} {Planning} {Methods} for {Autonomous} {Underwater} {Vehicle},'' \emph{Mathematical Problems in Engineering}, vol. 2021, p. 8847863, Feb. 2021, publisher: Hindawi. [Online]. Available: \url{https://doi.org/10.1155/2021/8847863}
\BIBentrySTDinterwordspacing

\bibitem{Yao_Review_2019}
\BIBentryALTinterwordspacing
T.~Yao, T.~He, W.~Zhao, and A.~Y. M.Sani, ``Review of path planning for autonomous underwater vehicles,'' in \emph{Proceedings of the 2019 International Conference on Robotics, Intelligent Control and Artificial Intelligence}, ser. RICAI '19.\hskip 1em plus 0.5em minus 0.4em\relax New York, NY, USA: Association for Computing Machinery, 2019, p. 482–487. [Online]. Available: \url{https://doi.org/10.1145/3366194.3366280}
\BIBentrySTDinterwordspacing

\bibitem{ScharffMarketReady}
J.~S. Willners, I.~Carlucho, S.~Katagiri, C.~Lemoine, J.~Roe, D.~Stephens, T.~Luczynski, S.~Xu, Y.~Carreno, E.~Pairet, C.~Barbalata, Y.~Petillot, and S.~Wang, ``From market-ready rovs to low-cost auvs,'' in \emph{OCEANS 2021: San Diego – Porto}, 2021, pp. 1--7.

\bibitem{campos}
C.~Campos, R.~Elvira, J.~J.~G. Rodríguez, J.~M. M.~Montiel, and J.~D.~Tardós, ``Orb-slam3: An accurate open-source library for visual, visual–inertial, and multimap slam,'' \emph{IEEE Transactions on Robotics}, vol.~37, no.~6, pp. 1874--1890, 2021.

\bibitem{octomap}
K.~M. Wurm, A.~Hornung, M.~Bennewitz, C.~Stachniss, and W.~Burgard, ``Octomap: A probabilistic, flexible, and compact 3d map representation for robotic systems,'' in \emph{Proc. of the ICRA 2010 workshop on best practice in 3D perception and modeling for mobile manipulation}, vol.~2, 2010, p.~3.

\bibitem{ompl}
I.~A. Sucan, M.~Moll, and L.~E. Kavraki, ``The open motion planning library,'' \emph{IEEE Robotics \& Automation Magazine}, vol.~19, no.~4, pp. 72--82, 2012.

\bibitem{fcl}
J.~Pan, S.~Chitta, and D.~Manocha, ``Fcl: A general purpose library for collision and proximity queries,'' in \emph{2012 IEEE International Conference on Robotics and Automation}, 2012, pp. 3859--3866.

\bibitem{chaudhary}
A.~Chaudhary, R.~Mishra, B.~Kalyan, and M.~Chitre, ``Development of an underwater simulator using unity3d and robot operating system,'' in \emph{OCEANS 2021: San Diego – Porto}, 2021, pp. 1--7.

\bibitem{BONINFONT}
F.~Bonin-Font, M.~{Massot Campos}, and A.~B. Burguera, ``Arsea: A virtual reality subsea exploration assistant,'' \emph{IFAC-PapersOnLine}, vol.~51, no.~29, pp. 26--31, 2018.

\bibitem{xia_visual-haptic_2023}
\BIBentryALTinterwordspacing
P.~Xia, H.~You, and J.~Du, ``Visual-haptic feedback for {ROV} subsea navigation control,'' \emph{Automation in Construction}, vol. 154, p. 104987, Oct. 2023. [Online]. Available: \url{https://www.sciencedirect.com/science/article/pii/S0926580523002479}
\BIBentrySTDinterwordspacing

\bibitem{garcia_natural_2017}
\BIBentryALTinterwordspacing
J.~C. García, B.~Patrão, L.~Almeida, J.~Pérez, P.~Menezes, J.~Dias, and P.~J. Sanz, ``A {Natural} {Interface} for {Remote} {Operation} of {Underwater} {Robots},'' \emph{IEEE Computer Graphics and Applications}, vol.~37, no.~1, pp. 34--43, Jan. 2017. [Online]. Available: \url{https://ieeexplore.ieee.org/document/7325202}
\BIBentrySTDinterwordspacing

\bibitem{louca_elicitation_2023}
\BIBentryALTinterwordspacing
J.~Louca, J.~Vrublevskis, K.~Eder, and A.~Tzemanaki, ``Elicitation of trustworthiness requirements for highly dexterous teleoperation systems with signal latency,'' \emph{Frontiers in Neurorobotics}, vol.~17, 2023. [Online]. Available: \url{https://www.frontiersin.org/articles/10.3389/fnbot.2023.1187264}
\BIBentrySTDinterwordspacing

\bibitem{teigland_operator_2020}
\BIBentryALTinterwordspacing
H.~Teigland, V.~Hassani, and M.~T. Møller, ``Operator focused automation of {ROV} operations,'' in \emph{2020 {IEEE}/{OES} {Autonomous} {Underwater} {Vehicles} {Symposium} ({AUV})}, Sep. 2020, pp. 1--7, iSSN: 2377-6536. [Online]. Available: \url{https://ieeexplore.ieee.org/abstract/document/9267917}
\BIBentrySTDinterwordspacing

\end{thebibliography}

\end{document}